# BRACU Mongol Tori: Next Generation Mars Exploration Rover

Niaz Sharif Shourov, Masnur Rahman, Mohammad Zahirul Islam, Ali Ahsan, Syed Md Kamruzzaman, Saifur Rahman, Md Sakiluzzaman, Intisar Hasnain, Ekhwan Islam, Saiful Islam, Md. Khalilur Rhaman
School of Engineering and Computer Science
BRAC University
Dhaka, Bangladesh

*Abstract*—**BRAC University (BRACU) has participated in the University Rover Challenge (URC), a robotics competition for university level students organized by the Mars Society to design and build a rover that would be of use to early explorers on Mars. BRACU has designed and developed a full functional next-generation mars rover, Mongol Tori, which can be operated in the extreme, hostile condition expected in planet Mars. Not only has Mongol Tori embedded with both autonomous and manual controlled features to functionalize, it can also capable of conducting scientific tasks to identify the characteristics of soils and weathering in the mars environment**

## I. INTRODUCTION

With the successful landing on planet mars in 1997 through NASA's pathfinder mission, research regarding mars exploration rovers has been increased among the student level. As a result competition like University Rover Challenge (URC) has been introduced in 2006 by The Mars Society for the university students aims to design and build the next generation of mars rover. As every space agency is planning to have a human mission in mars so the need of assistant rovers which can work along with the astronauts are pretty much high.

Since 2007 a lot of experimental research has been conducted by several students all over the world on planetary rovers for mars exploration. With the help of Rover Challenge Series such as University Rover Challenge URC), European Rover Challenge (ERC), Canadian International Rover Challenge (CIRC) and Indian Rover Challenge (IRC) students can test their mars rovers figuratively and relate self-maid custom technologies on the simulated Martian terrain created by the competition organizers. In this following research a six wheeled next generation mars exploration rover developed by a student team named "BRACU Mongol-Tori" from BRAC University, Bangladesh has been demonstrated and presented successfully. This paper focuses on adding and testing innovative approaches in mars rover, e.g., design and development of chassis, suspension system, robotic arm, custom wheels. Also, this paper considers custom made control software & circuit, long range communication system, autonomous navigation, sample cache collection and both onboard, laboratory analysis.

The developed rover can successfully perform several tasks. Traverse through rough and sandy terrain, assistance task like turn on/off switch, turning knobs, carrying tools, autonomous traversing based on GPS location and soil sample collection with onboard analysis are some examples of the completed tasks. This paper will describe each and every subsystems with thorough investigation of development process.

## II. SYSTEM ANALYSIS

For developing a mars rover, a multidisciplinary team is needed for different purpose. In this research a team of 20 undergraduate students from different departments like Computer Science & Engineering, Electrical & Electronic Engineering, Electrical & Communication Engineering, Business Administration, and Pharmacy has worked together to maximize the result. The research team has been divided into five sub teams and worked with the following subsystems (a) mechanical (b) electronics (c) communication (d) software (e) science.

### A. Mechanical

The mechanical subsystem has been divided into couple of modules. They are as follows:

**1. Chassis:** The chassis of the rover is stimulated from the combination of ladder and tube space frame design. Triangular shaped chassis has a square box 0.635*0.254 meter at the front. To increase rigidity and stiffness as well as to diminish the possibility of shape distortion due to the application of bumpy pressure triangulation has been incorporated in the chassis. Several research shows that triangular shape structure has more strength than a conventional square shape structure. Flexing loads are transmitted as tension and compression loads along the length of each triangle strut shown in Fig. 1.

**2. Suspension:** Considering the rough surface and the vertical drops a modified semi rocker bogie suspension system has been incorporated in the system. The design is inspired by the NASA's popular rocker-bogie suspension system. In this following design, the bogies are mounted on to the front part of chassis and the rear suspension is supported by a high-performance shock absorber with double spring. Some existing systems uses V shaped structures for the front bogie, but in the following system U shaped structure has been implemented for the bogies.

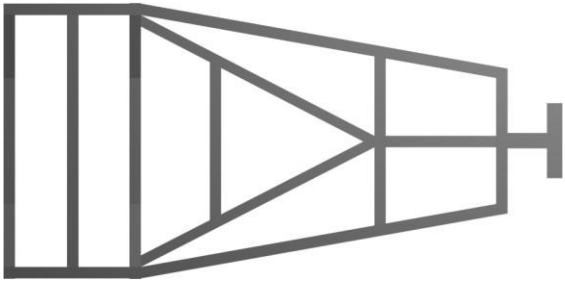

Fig. 1. Chassis

For this U shape the system achieved maximum angle between the front two wheels of one side which helps the system to successfully performing the vertical drops.

**3. Steering:** For rotating the rover easily without dragging an independent steering mechanism has been installed at the front two wheels of the system. Two linear actuators has been incorporated for controlling the steering system. The system can rotate the front wheels 35 degree in both left and right direction.

**4. Wheel:** After several experiments with different size of wheels the optimal size for required martian surface is 0.3048 meter diameter with a width of 0.127 meter. Considering the light weight and high strength requirements molded aluminum has been selected as a raw material for the wheels. For avoiding the collision with rocks motors are placed inside the wheels. External rubber grip on the wheels has been incorporated with a unique orientation. The orientation of the grip helps the rover to rotate 360 degrees, climbing the rock and deal with the vertical drops.

**5. Robotic Arm:** A very strong and efficient robotic arm with six degrees of freedom (base, shoulder, elbow, wrist, two degrees of freedom in the end effector) is designed, which is shown in Fig. 2. The arm consists of three linear actuators and three dc motor. A gear mechanism has been incorporated at the base of the arm for the 360-degree rotation. The three-fingered end effector with rubber in the fingers can grab small stuff perfectly. The end effector is portable that's why different type of gripper or module can be added with the robotic arm with in no time. An intelligent rotational gear based system has been comprised of the claw for 360-degree rotation without wire twisting. The claw can hold and the arm can lift up to 5 kg of weight under any circumstances.

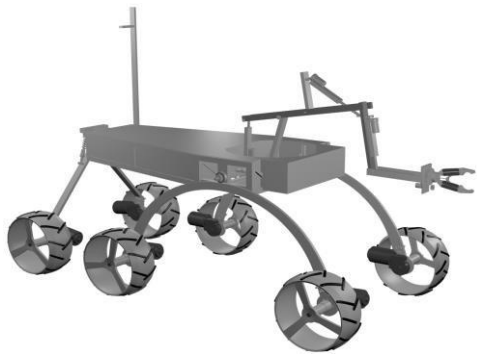

Fig. 2. 3D design of Rover

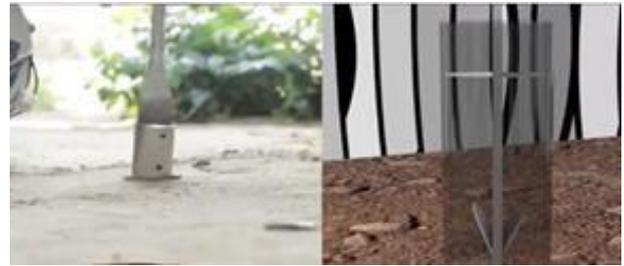

Fig. 2. Sample Collector

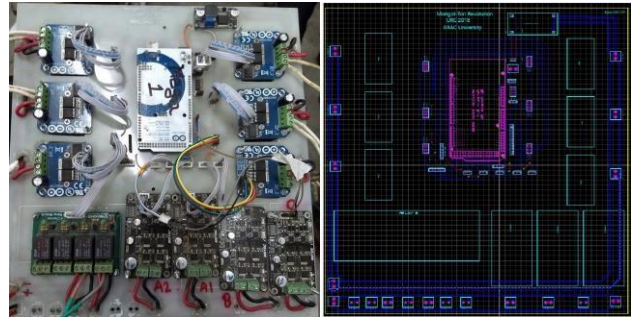

Fig. 4. Circuit Diagram

**6. Sample Cache Collector:** After couple of experiments a sophisticated drill mechanism has been incorporated with the robotic arm to collect subsurface soil from more than 10 cm depth. This drill is not only creating 10 cm deep hole but also collect soil layer by layer with the help of its unique hinge door mechanism which was installed at the front part of the drill module shown in Fig. 3 so that it can be used for lab test later on.

*B. Electronics*

For an efficient power distribution and electronics system the total circuit system has been divided into three sections. The sections are as follows:

**1. Control Circuit:** The main goal of this section is to control the wheels and the robotic arm. As the system consists of six wheels so there are six different high torque dc motors has been integrated in the system. After several experiments the maximum current was measured with full load from the wheel motor was approximately between 10A-12A. Depending on the motor current consumption BTS 7960 has been selected as motor driver for its high current rating which is 43A. It contains one p-channel high side MOSFET and one n-channel low side MOSFET connected with an integrated driver IC which create high current half bridge. The driver has PWM features which can be operated up to 25 kHz. Separate PCB has been designed for wire connection reduction between the microcontroller unit (MCU) and the motor driver units. The robotic arm of the system consists of 3 linear actuators and 3 dc motors. The no load current consumption of the arm is 0.7A to 1.2A. With a load of 5 kilogram the maximum current measured in the following robotic arm is between 5A – 6A. The arm control PCB is consist of 6 Cytron MD10C motor driver module which is a fully NMOS H-bridge driver. This bidirectional controller works between 5V-25V and a maximum current of 13A continues with 30A peak for10 seconds. The PWM feature can be applied up to 20 KHz in this driver. Both the wheel and arm control PCB are connected with the master

MCU where the MCU is connected with rover onboard computer Intel NUC shown in Fig. 4.

**2. Sensors:** In this section a sensor fusion box has been developed for the rover. The fusion is consist of seven different sensors which can collect data from the weather as well as from the soil. MQ-135 sensor is responsible for measuring $CO_2$ while MQ-7 can help the system to collect CO amount in the air. Temperature and humidity data is achieved by the DHT-11 sensor. For the stability test of the rover chassis MPU-6050 sensor has been incorporated in the system which has built in accelerometer and gyro chip. DS18B20 sensor has been installed into the fusion for the soil temperature data. For finding the amount of water in the soil a soil moisture sensor is added into the sensor fusion. A separate dedicated PCB in Fig. 5 has been designed for an efficient plug and play connection.

**3. Power Distribution:** For reducing the load from a single power source multiple power source has been incorporated in the system for powering up different modules. A custom made power distribution board control all the power distribution with proper safety. The distribution system is divided into three sections. For producing the power to section-1 which consist of motor drivers, motors and actuators two lithium polymer (LiPO) batteries with 10000 mAh 11.1V rating has been incorporated in the system in series connection which increase the voltage into 22.2V. Section-2 is consist of Intel NUC on board computer and sensor fusion box. A 5400 mAh 11.1V LiPO battery has been dedicated for producing 12V to NUC pc and 5V to sensor box. The section-3 is comprised with the communication module and camera module where communication module consume 24V and camera module consume 12V. Two LiPO batteries with 5400 mAh rating has been used in series connection for 24V output which produce sufficient amount of power to the section-3. XL6009 dc-dc converter converts voltage from 24V to 12V & 5V. Switching frequency of this module is near 400 kHz. This converter module is installed into the main power distribution board. For safety of the total electronics segment extra cooling system is developed which includes two DC fan and aluminum heat sink on the regulator IC. Moreover a kill switch has been integrated for any emergency situation.

*C. Communication*

**1. Base Station:** To ensure a strong communication link between base station and rover, p2p (point to point) connection between the routers of rover and base station has been set up. After multiple test run, link budgeting and optimizing the setup 2.4 GHz band is used for control end. Different outdoor router pairs were tested to get the maximum range and latency of the command. For testing maximum coverage ping commands have been sent from one pc to another while increasing distance gradually. Variation of directionality and SNR were observed for different distant location. Based on which Mongol tori team finalized 2.4GHz Ubiquiti Airmax Rocket M2 [8], which is compatible with both omnidirectional and sector antenna of the three actuators and three DC motors needs 12V which have been managed directly from the battery source through motor drivers. Though motor drivers have some built-in protection for keeping IC cool, extra cooling system is developed which includes two DC fan and aluminum heatsink on the regulator IC. In spite of having all these safeties, a kill switch has been integrated for any emergency situation. It was set as Access point bridge mode at the base station. To ensure clear line of sight (LOS) communication, signal distortion can be kept the minimum and for this purpose, focused narrow beam signal having significant directionality must be transmitted from the transmission side. This made the team to choose airMax sector antenna of 120-degree directional coverage [9]. Control data requires better reliability hence TCP protocol is used for data transmission.

**2. Rover setup:** At rover 2.4GHz Ubiquiti Bullet M2 was used which was set as station mode. To extend horizontal beam pattern coverage, external omnidirectional antenna, TL-ANT2412D [10] was connected to the router. Angular coverage needs to be maximum at the rover. Since it is subjected to motion and needs to carry out the task in any direction assigned to. It is required to manipulate control data received at any instantaneous position. A Separate wireless device of 5.8GHz Band having 48 channel (TS832 and RC832) [11] was used for transmission and reception of low composite video feed from cameras installed in rover with low latency. This setup enhanced flexibility by reducing the dependence on the 2.4GHz router for video. Team Mongol Tori decided to use two different wireless bands, as it results in no interference between the control signal and video stream. UDP protocol allows faster transmission, with delays as minimal as possible, to capture video stream this particular protocol have been handy. FPV camera pair having 720P resolution and viewing angle of 165 degrees have been used in the rover, to capture and transmit the live video stream.

*D. Software*

The rover software establish a central control system from a remote distance which consists of control interface at base station and central rover controller. Each system is responsible for controlling different aspects of the rover. The base station is responsible for interpreting data from the controller, processing data received from the rover and sent data to the rover. This systems' main aspect is to provide reliable, easy-to-use control of the rover, accurate and up-to-date information of the rovers' status. The mapping GUI helps to track the rover in real time on a pre-loaded map using the latitude and longitude gotten form the GPS sensor. ATmega2560 based Arduino microcontroller enables smooth control over the motors. The microcontroller is directly connected with the on board NUC PC.

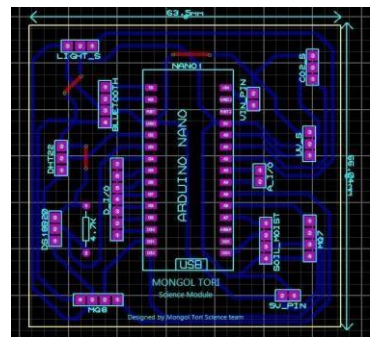

Fig. 5. Sensor Module PCB

**1. Control GUI:** For controlling the rover movement a control GUI is developed using Java programming language as shown in Fig. 5. In this system the base station acts as the client while the rover acts as the server and the protocol followed in the communication is Transmission Control Protocol (TCP). The base station software takes command from the keyboard according to movement, process the command and send an instruction to rover side software which runs on the NUC pc. The rover side software interprets the command and runs the rover accordingly.

**2. Offline Map:** To locate and plot the rover's position, Grove GPS module is being used which features 22 tracks / 66 acquisition channel GPS receiver. GPS data is received in NMEA format and every received string is sent to the base station. The software in the base station is developed in python 2.7 which processes the raw GPS data and extracts the latitude and longitude of the rover position and continuously plots them in the cropped picture of the static map using matplotlib and pynmea.

**3. Stability GUI:** An MPU-92/65 sensor is used which is a single chip package containing a 3 axis accelerometer, a 3 axis gyroscope, a 3 axis compass sensor. The sensor is connected to ATmega328P based microcontroller Arduino Uno which communicate with the base station GUI to provide a clear and accurate orientation of the rover. The stability GUI is developed using Arduino and processing.

**4. Sensor GUI:** The science GUI developed in C# interprets the sensor readings into custom gauges which offers a graphical perception that makes it easy to distinguish between different states of sensor readings. To clearly present the data, changes various dials, meters and gauges are used in the sensor GUI as shown in Fig. 6

*E. Science*

Mars is the fourth plane of this solar system located around 142 million [1] miles away from the sun. It is one of the four rocky planets that have an almost same duration of night and day like earth but it has only 37.5% gravity of earth so the atmosphere of Mars is thin and mostly contains carbon dioxide and some water vapor [1]. Due to this thin atmosphere and lack of magnetic field, the planet is defenseless against radiation. The surface of this planet is exposed to solar energetic particles and galactic cosmic rays at the rate of 0.67 millisieverts per day [2]. Unlike Earth, most of the Martian surface is covered by the mantle.

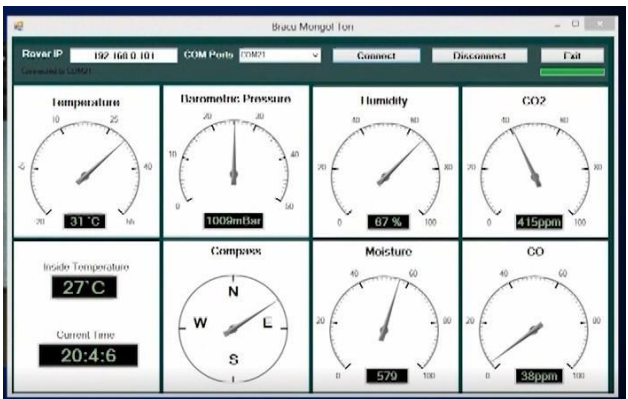

Fig. 6. Sensor GUI

On this rover "Mongol Tori", on board instruments are used to examine the environmental condition to support life and some quick lab tests are designed to get the condition of the soil. Firstly, there is a digital temperature and humidity sensor on board for measuring air condition which contains a capacitive humidity sensor and a thermistor. High humidity with the combination of aerosols (small particulates) helps the formation of the droplet and increase the possibility of raining and water cycle [3]. Different gas sensors have been used to measure carbon dioxide and carbon monoxide concentration in ppm which is also indicate the survivability of life forms. Additionally, a sophisticated drill mechanism is developed to collect subsurface soil from more than 10 cm deep. This drill is not just only creating 10 cm deep hole but also collect soil layer by layer so that it can be used for lab test later on.

This rover also has Grove moisture sensor and Waterproof Digital Temperature sensor to measure the moisture and temperature of the soil. As the air pressure of Mars is much lower, the possibility of finding water is high in subsurface than the top surface. Secondly, after collecting the soil from drill cylinder we can take the layer from five cm to ten cm range and operate some quick lab test. Analog pH meter is used to measure the pH of the soil. Most of the life form on earth found within the pH range of 6.5-9.0 [5]. On the other hand, mars have 8-9. Another Important two tests the team has conducted is biomass test and water capillary test.

In the biomass test, the soil have been heated up for few minutes and the mass difference is calculated to get the idea of water and other liquid bio-substance present in the soil. Furthermore, the water capillary test indicates the water absorbing capacity of the soil. This is important because if the water holding capacity is low the rainwater goes down deep under the ground and surface remains dry. So, the sample collecting system and series of lab test will give a clear idea of the environment there to support life within the short amount of time.

### III. AUTONOMOUS TRAVERSAL

For the implementation of the autonomous traversal, BRACU Mongol-tori used U-blox 6 based NEO-6 series GPS module, an HMC5883L three-axis magnetic module and computer vision. GPS module is used to calculate the latitude and longitude with a 3 meter error margin which is experimented in open environment with preloaded map. The Compass sensor is used along with the GPS module to find out the exact rover direction. Several target gates are selected to get the expected traversal path. The image processing sub-routine takes over once the rover comes within 3.5 meter radius of the destination dictated by the GPS values. Within this radius the autonomy of the rover is maintained by the image processing routine constructed using OpenCV library and python programming language. After getting the values from the GPS sensor the software calculates the distance and the angle between start and every end positions which were calculated using the Haversine formula. Evaluating the direction data from the magnetometer it aligns the rover in the direction of the determined travel path. Once within range of the destination the software switches to image processing procedure based on the feedback from the on board mcu for autonomous traversal.

### IV. RESULTS

For verifying a system there is no other better option than real life experiment. For the following system that has been developed by summing up couple of sub systems operational testing is the correct way to finding the durability and productivity for the practical scenario. The total testing has been divided into three phases. They are as follows:

*A. Sub – System Testing*

There are couple of sub – systems which has been developed for fulfilling the total systems criteria. The sub – systems are wheel, suspension and chassis, end effector, sample collector and software. For testing the wheels a custom made ramp has been used where one single wheel can be fixed with a motor connected to it and the wheel can move linearly on the ramp. Different types of obstacles has been added on the route of the ramp. After power up the motor the wheel overcome maximum number of obstacles depending on different size. The wheels can overcome rocks with the length of 0.254 meter which was placed with 90 degrees of angle. After testing the wheels the chassis and suspension parts are also been tested in an indoor area. The stairs, different types of objects and volunteer team member's body has been used as obstacle for the suspension & chassis test. Stairs with the height of 0.2032 meter has been easily overcome by the rover. Obstacles like a wooden box with a height of 0.2794 meter and 80 degrees angle has been easily overcome by the rover. While overcoming an obstacle by a single wheel the other six wheels of the rover remain grounded which specify a successful suspension system. For testing the end effector different types of materials has been selected for grabbing and uplifting purpose. The thinnest material that has been successfully grabbed by the following end effector was a round shape stainless steel bar with a diameter of 0.00762 meter. The thickest material has been grabbed successfully by the end effector was a 0.1016 * 0.1016 meter metal steel box pipe. For testing the sample cache collector a tub has been used with full of soil. The cache collector can successfully dig up to 0.1016 meter inside the soil and collected soil sample from that deep portion. For controlling the rover a platform independent graphical user interface (GUI) has been developed which can be used for controlling any type of system or robot. For testing the fluentness of the command GUI a dummy robot has been used which was a four wheeler fighting robot. The lag between the command and the execution was 3 to 7 sec depending on the distance.

*B. Complete System Testing*

For complete system testing, communication and electronics system, complete robotic arm are considered. For testing the communication system an open field has been selected with line of sight. The rover crosses approximately 900 meter of distance with full strength communication network. After crossing the approximate distance of 1050 meter the communication system doesn't work properly. During the communication range test the rover driving circuit was also been tested in extreme hot weather. The temperature was 38 degree Celsius during the testing. Moreover all the circuit modules works properly and handled all the ampere generated from the motor and actuators due to rough terrain. For testing the complete robotic arm different types of material has been selected for uplifting. Two separate bottle full of water has been set for uplifting. Bottle-1 has a weight of 5 kg and bottle-2 has a weight of 6 kg. The robotic arm successfully uplift the bottle-1, but while uplifting the bottle-2 one of the finger has broken and the system cannot uplift the bottle-2.

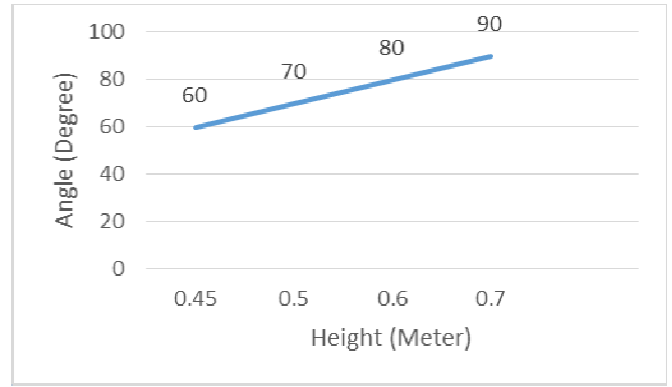

Fig. 7 Angle vs Height Graph for Vertical Drop

*C. Operational Testing*

For operational testing the rover was fully equipped with all the system and sub – system. Three types of task has been selected for testing the rover's task fulfill based efficiency. The three task are as follows: Extreme Terrain Travers, Equipment Servicing and Autonomous Traversal. The extreme terrain traversal task is basically the rover needs to traverse through extremely uneven terrain which include vertical drops, steep slope, and different types of obstacles. The following developed rover faces different types of vertical drops from different types of height.

In Fig. 7, the X axis is considered as Height and the Y axis is considered as Angle. So when the rover overcome a vertical drop with an angle of 60 degree then the height of the vertical drop was at 0.45 meter. In this way when the angle of the vertical drop was 90 degree then the maximum height overcome by the developed rover was 0.7 meter. After that when the rover attempts vertical drop for the height of 0.8 meter with an angle of 90 degree then the rover face an upside down situation. The following system can successfully overcome steep slope with the maximum angle of 35 degree which has a maximum height of 1.2 meter. The equipment servicing task include pouring liquid into a specified hole through a bottle, turning on any types of switch etc. During the experiment the rover successfully pour liquid into a hole which was at a height of 0.6 meter. The rover successfully turn on and off different types of toggle and push button switch using its efficient end effector. For testing the autonomous traversal task six different GPS coordinates has been note downed which are situated at different distance from each other. The rover successfully gone through the first three coordinates autonomously without any manual command. The first three coordinates were set at an approximate distance of 10 meter from each other. The fourth coordinate were set at an approximate distance of 20 meter. While crossing the fourth coordinate the rover move from the correct path which will lead the rover to the exact fourth coordinate. With a distance of 4 meter from the exact coordinate the rover cross the fourth coordinate.

V. CONCLUSION

With a vision of colonize the red planet all the technologies of 21st century are working together. For developing a colony on planet mars the gadget mars rover can play a vital role. In this research paper we have tried to present step by step process of developing the next generations of mars rovers. The main focus of this research was to develop a rover from scratch. Moreover development of the chassis with a new formation and adding steering into the system is a significant contribution of the research. Communicating with the rover with a distance up to 1000 meter without any internet connection is also a significant milestone. Developing a universal platform independent control GUI which can be used in any type of control system and generating an offline map is also a notable contribution to the research. Moreover from the collection of soil sample different types of data has been acquired through sensors and different types of scientific analysis and experiments. Autonomous traversal was a remarkable attempt of this research, though it was partially successful. For the future development of this research many upgrade can be added like autonomous traversal mood can be modified to achieving the 100% autonomy rate. Furthermore the scientific analysis of the soil sample can be done fully onboard where external experiment will not be required anymore for finding the important data regarding soil sample. This may be help the system to be self-depended.